\documentclass{article}

\usepackage[preprint]{neurips_2024}
\usepackage{usebib}

\usepackage[utf8]{inputenc} 
\usepackage[T1]{fontenc}    
\usepackage{hyperref}       
\usepackage{url}            
\usepackage{booktabs}       
\usepackage{amsfonts}       
\usepackage{nicefrac}       
\usepackage{microtype}      
\usepackage{commands}
\usepackage{bm}
\usepackage{pgfplots}
\pgfplotsset{compat=1.18}

\usepackage{tikz}
\usepackage{standalone}
\usepackage{enumerate}
\usetikzlibrary{positioning}
\usepackage[dvipsnames]{xcolor}
\usepackage{blindtext}
\usepackage{textcomp}
\usepackage{pgfplots}
\usepgfplotslibrary{fillbetween}
\usepackage{multirow}

\usepgfplotslibrary{dateplot}
\usepackage{caption}
\usepackage{subcaption}
\usepackage{algorithm} 
\usepackage[noend]{algpseudocode}
\usepackage{xcolor,colortbl}
\usepackage{hhline}
\title{Self Paced Gaussian  Contextual Reinforcement Learning}

\author{%
  Mohsen Sahraei Ardakani\\
 Department of Electrical and Computer Engineering \\
  North Carolina State University\\
  Raleigh, NC\\
  \texttt{msahrae@ncsu.edu} \\
  \And
  Rui Song \\
  Department of Statistics \\
  North Carolina State University\\
  Raleigh, NC\\
  \texttt{rsong@ncsu.edu} \\
}

\begin{document}

\maketitle

\begin{abstract}
Curriculum learning improves reinforcement learning (RL) efficiency by sequencing tasks from simple to complex. However, many self-paced curriculum methods rely on computationally expensive inner-loop optimizations, limiting their scalability in high-dimensional context spaces. In this paper, we propose Self-Paced Gaussian Curriculum Learning (SPGL), a novel approach that avoids costly numerical procedures by leveraging a closed-form update rule for Gaussian context distributions. SPGL maintains the sample efficiency and adaptability of traditional self-paced methods while substantially reducing computational overhead. We provide theoretical guarantees on convergence and validate our method across several contextual RL benchmarks, including the Point Mass, Lunar Lander, and Ball Catching environments. Experimental results show that SPGL matches or outperforms existing curriculum methods, especially in hidden context scenarios, and achieves more stable context distribution convergence. Our method offers a scalable, principled alternative for curriculum generation in challenging continuous and partially observable domains.
\end{abstract}

\section{Introduction}
Reinforcement learning (RL) provides a general framework for sequential decision making where an agent learns a policy through trial-and-error interaction with an environment in order to maximize cumulative reward \citep{Sutton2018,kaelbling1996reinforcement}. The combination of RL with deep function approximation has led to impressive results in domains such as games and robotics, but these successes often rely on massive amounts of experience, dense or carefully shaped rewards, and extensive hyperparameter tuning. In many practical problems, agents face sparse and delayed rewards, high-dimensional state and context spaces, and substantial environment stochasticity, which make naive exploration and monolithic training procedures sample-inefficient and brittle.

Curriculum Learning (CL) was originally introduced in supervised learning as a strategy to improve optimization and generalization by exposing a learner to examples in a meaningful order, typically from simple to complex \citep{Bengio2009}. Rather than modifying the objective function, CL shapes the training distribution, smoothing the optimization landscape and guiding the learner away from poor local minima. Subsequent work has formalized and extended this idea through self-paced learning and related variants that automatically adapt the training distribution based on the learner's progress \citep{kumar2010self,jiang2015selfpaced,jiang2014selfdiversity,ren2018self}. These methods construct a curriculum not by manually specifying difficulty, but by solving an auxiliary optimization problem that reweights or selects samples according to evolving notions of easiness, informativeness, or diversity.

Curriculum Reinforcement Learning (CRL) adapts these principles to sequential decision making problems \citep{portelas2020automatic}. Instead of training a policy from scratch on a fixed task distribution, CRL organizes experience as a sequence of tasks, environments, or initial conditions whose difficulty gradually increases. This perspective is particularly natural in contextual Markov decision processes (CMDPs) \citep{hallak2015contextual}, where each task instance is parameterized by a context vector and the learner is evaluated on its ability to generalize across contexts \citep{benjamins2021carl,benjamins2022contextualize, sahraei2025contextual}. Shaping the context distribution over the course of training can dramatically improve sample efficiency and robustness, but raises new algorithmic and computational design questions so that, the curriculum must remain adaptive, scalable to high-dimensional context spaces, and compatible with standard deep RL backends.

Self-paced curriculum reinforcement learning (SP-CRL) instantiates this idea by treating curriculum design as an optimization over a task or context distribution that trades off performance and diversity. Building on self-paced learning formulations from supervised learning \citep{kumar2010self,jiang2015selfpaced,jiang2014selfdiversity,ren2018self}, SP-CRL methods for CMDPs adapt the context distribution based on the agent’s current capabilities, gradually shifting probability mass from easy to more challenging contexts \citep{klink2020sp,klink2020spdrl,eimer2021self}. Empirically, such self-paced curricula have been shown to accelerate training and improve generalization across tasks and contexts \citep{portelas2020automatic,eimer2021self}. However, many of these methods rely on computationally intensive inner-loop optimizations over the context distribution, often implemented as iterative numerical or gradient-based solvers at each curriculum update \citep{klink2020spdrl,huang2022curriculum,klink2022constrained_ot}. This additional optimization can become the dominant cost, limiting the scalability of CRL in large-scale or real-time settings.

\textbf{Contribution.} In this work, we propose a CRL algorithm that is compatible with standard deep RL pipelines while avoiding expensive inner-loop numerical optimization over context distributions. Our method can be seamlessly combined with conventional deep RL algorithms, matches or exceeds the performance of state of the art CRL baselines on benchmark CMDPs, and substantially reduces computational overhead by replacing inner-loop distribution optimization with a more efficient update mechanism. As a result, our approach extends the practical applicability of curriculum learning to higher dimensional context spaces and more demanding real-world scenarios.

\section{Related Work}

Curriculum learning has been extensively investigated as a means to improve optimization and generalization in reinforcement learning by shaping the distribution of training tasks rather than directly modifying the underlying objective \citep{Bengio2009,portelas2020automatic}. Early work in Curriculum Reinforcement Learning (CRL) demonstrated that manually or heuristically organizing tasks from simple to complex can substantially accelerate learning in sparse-reward and hard-exploration problems. Examples include reverse curriculum generation from goal states, automatic goal generation, and environment generation mechanisms that progressively expand the region of the state space in which the agent is trained \citep{florensa2017reverse,florensa2018automatic,andrychowicz2017her,ghosh2017divide}. These methods provided compelling empirical evidence that a carefully designed exposure schedule over tasks enables agents to learn behaviors that would be prohibitively difficult to acquire from a static task distribution.

Subsequent research has moved from hand-crafted curricula towards automated curriculum mechanisms in which a teacher or scheduler selects tasks based on signals such as learning progress, accuracy, or coverage of the task space. Accuracy-based curricula and teacher algorithms exemplify this trend, automatically adapting the task distribution to the learner’s evolving competence while maintaining sufficient diversity \citep{fournier2018accuracy,portelas2019teacher,yengera2021curriculum_demo_teaching}. A broader line of work has extended these ideas to multi-task RL, meta-RL, and cooperative multi-agent settings, where curricula are used to orchestrate the exposure of agents to families of tasks, opponents, or interaction patterns \citep{liu2021curriculum_offline_imitation,chen2021variational,gur2021code_zero_shot_rl,feng2021neural_autocurricula,portelas2020automatic}. These contributions collectively show that curriculum mechanisms can be integrated with a wide range of base RL algorithms and problem domains, yielding consistent gains in sample efficiency and robustness.

Contextual Markov Decision Processes (CMDPs) provide a natural formalism for studying such curricula, as they explicitly parameterize tasks by context variables and assess performance in terms of generalization across contexts \citep{hallak2015contextual}. Benchmarks such as CARL highlight the importance of contextual variation for evaluating RL algorithms and curriculum strategies \citep{benjamins2021carl,benjamins2022contextualize}. Within this framework, curriculum learning corresponds to shaping a sequence of context distributions from which tasks are sampled. Self-paced and self-adaptive methods operationalize this idea by optimizing the context distribution itself. Self-paced contextual reinforcement learning and related variants define auxiliary objectives that trade off performance and regularization terms such as diversity or proximity to a target distribution, and update the curriculum by solving an inner optimization problem over contexts \citep{klink2020sp,klink2020spdrl}. Self-Paced Context Evaluation (SPaCE) follows a similar principle, using performance-based criteria to selectively emphasize contexts that are most informative for the current agent, and has been shown to improve sample efficiency and convergence behavior compared to round-robin baselines \citep{eimer2021self}. These approaches build on earlier self-paced learning formulations from supervised learning \citep{kumar2010self,jiang2015selfpaced,jiang2014selfdiversity,ren2018self}, extending them to sequential decision making and high-dimensional context spaces.

More recently, a number of works have proposed principled curriculum mechanisms grounded in optimal transport, domain adaptation, and distributional robustness. Optimal transport based CRL methods interpret curriculum construction as transporting probability mass from an auxiliary (easier) task distribution to the target task distribution under geometric or constraint based criteria \citep{huang2022curriculum,huang2022curriculum_ot_gradual,klink2022constrained_ot}. These methods provide finer control over how task difficulty evolves and can induce smooth curriculum trajectories between auxiliary and target domains. Other contributions investigate theoretical aspects and more complex curriculum architectures, including analyses of complexity gains induced by curricula in single task RL \citep{li2023understanding}, mixture of experts curricula for acquiring diverse skills \citep{celik2024acquiring}, proximal curricula that exploit task correlations \citep{tzannetos2024proximal}, and genetic or evolutionary curricula in multi agent environments \citep{song2025genetic}. While these approaches further expand the design space of curricula and demonstrate strong empirical and theoretical benefits, a common limitation is their reliance on computationally expensive inner-loop optimization over task or context distributions, teacher policies, or auxiliary models. This additional complexity can hinder scalability, especially in high dimensional CMDPs or real-time applications, and motivates the development of CRL algorithms that retain the benefits of adaptive curricula while reducing their computational overhead.



\section{Preliminaries}
Contextual Markov Decision Processes (CMDPs) enhance traditional Markov Decision Processes (MDPs) by incorporating context into decision-making, allowing for a dynamic adaptation based on environmental changes or task-specific details. The CMDP framework is represented by a tuple $(\CS, \SS, \AS, \MS(c))$ where $\CS$ specifies the context space, and similar to MDP definition $\SS$ and $\AS$ are state and action spaces. Lastly, $\MS$ is a function mapping from any context $c \in \CS$ to an MDP $\MS(c) = (\SS, \AS, P^c, r^c, \rho^c)$. CMDP can be regarded as a set of MDPs sharing the same state and action spaces.   
It should be noted that during one episode the context assumed to be fixed. Depending on the definition of the context variable  the nature of the problem changes and requires different learning algorithms. In the literature three main classes of problems are explored: 1- discrete finite and observable  context set\cite{eimer2021self}, 2- continuous and observable context set \cite{klink2020spdrl}, and 3-  non-observable context\cite{rakelly2019efficient}.

The objective in a CMDP is to determine a policy \( \pi: S \times C \to \Delta(A) \) that maximizes the expected cumulative reward over trajectories $\tau = \{(s_t, a_t)|t\geq 0\}$ over context distribution $\psi(c)$ with a discount factor of $\gamma\in [0,1)$, hence the objective function would be:
\begin{equation}
J(\psi, \pi) = \EXP_{\psi(c), p_\pi(\tau|c)}\left[\sum_{t\geq 0}\gamma^t r_c(s_t, a_t)\right],\quad p_{\pi}(\tau|c) = p_{0,c}(s_0)\prod_{t\geq 0} p_c(s_{t+1}|s_t,a_t)\pi(a_t|s_t, c)
\end{equation}

\section{Self Paced Gaussian Curriculum Learning}
It has been shown that not only the context distribution substantially contributes to the learning efficiency but also changing it during the training can be an effective tool for the agent to learn challenging tasks \cite{klink2020spdrl,klink2020sp,benjamins2021carl,eimer2021self}. Self Paced Curriculum Learning aims at learning a context distribution during the training which makes the learning efficient and converges to the target distribution at the end. Assuming the context distribution $p_\nu(c)$ is parametrized by $\nu \in \R^m$, the RL objective can be reformulated as 
\begin{equation}
    \max_{\nu, \omega} J(\nu, \omega) - \alpha D_{\text{KL}}\left( \psi(c)\parallel p_\nu(c) \right), \quad \alpha \geq0. \label{eq:loss}
\end{equation}

Where $D_{\text{KL}}( \psi(c)\Vert p_\nu(c))$ represents the KL divergence between the agent's sampling context distribution and the target context distribution, and $\alpha$ going from $0\to \infty$ regularizes the optimization such that the sampling distribution enables higher returns at the beginning and is pushed towards the target context distribution at the end.

\begin{align}
\mathcal{D}_i =&\left\{(c^k, \tau^k)|k\in[1, K], c_k\sim p_{\nu_i}(c), \tau_k \sim \pi_{\omega_i}(\tau|c_k)\right\}\label{eq:traj}\\
    &\max_{\nu_{i+1}} \frac{1}{K}\sum_{k=1}^{K}  \frac{p_{\nu_{i+1}}(c^k)}{p_{\nu_i}(c^k)}V_{\omega_i}(s_0^k, c^k) - \alpha_i D_{\text{KL}}\left( \psi(c) \parallel p_{\nu_{i+1}}(c)\right) \label{eq:ctx_op}\\ \text{s.t. }& D_{\text{KL}}\left(p_{\nu_{i+1}}(c)\parallel p_{\nu_{i}}(c)\right) \leq \epsilon\\
\nonumber\\
       \ & \min_{\nu_{i+1}} \ D_{\text{KL}}\left(\psi(c) \parallel p_{\nu_{i+1}}(c)\right)\label{eq:perf_op}\\
    \text{s.t. } & \frac{1}{K}\sum_{k=1}^{K} \frac{p_{\nu_{i+1}}(c^k)}{p_{\nu_i}(c^k)}V_{\omega_i}(s_0^k, c^k) \geq V_{-}\label{eq:perf_cond}\\
    & D_{\text{KL}}\left(p_{\nu_{i+1}}(c)\parallel p_{\nu_i}(c)\right)\leq \epsilon
\end{align}

Adopting the results of \citep{klink2020sp}, the context distribution update in the  RL optimization Objective~\ref{eq:loss} can be separated from the policy updates, and stated as Objective~\ref{eq:ctx_op}. Furthermore, it has been proven  for a given set of collected trajectories, $\mathcal{D_i}$,  the optimization problem is reduced to two subproblems without the need for scheduling $\alpha$\citep{klink2020spdrl}; If the minimum performance condition in Inequality~\ref{eq:perf_cond}, is not met, the agent solves the performance optimization problem of Objective~\ref{eq:perf_op} and  learns to make the contexts with higher expected value (i.e. the easier contexts) more probable. Afterwards, the agent solves the context convergence problem defined by Objective~\ref{eq:cnvg_opt}, which gradually shifts the sampling context distribution towards the target context distribution.
\begin{align}
   & \min_{\nu_{i+1}} \ -\frac{1}{K}\sum_{k=1}^{K}\frac{p_{\nu_{i+1}}}{p_{\nu_i}}V_{\omega_i}(s_0^k, c^k) \label{eq:cnvg_opt}\\
    \text{s.t. } & D_{\text{KL}}\left(p_{\nu_{i+1}}(c)\parallel p_{\nu_i}(c)\right)\leq \epsilon
\end{align}

\subsection{Performance Optimization Problem}
The  context distribution is specified as a multivariate normal distribution with target mean $\widetilde{\mu}$ and target covariance matrix $\widetilde{\Sigma}$. Then one, parameterization of variational context distribution is $\mathcal{N}(\mu, \Sigma = \Theta^{1/2}\widetilde{\Sigma}\Theta^{1/2})$ leading to  $\nu = [\mu, \bm{\theta}]\in \R^{2d}$ where $\Theta = \mathrm{diag}( \bm{\theta})$ is a diagonal matrix scaling the covariance of the context distribution and $d$ is the dimension of the context space. The KL-divergence for multivariate normal distribution has a closed form, then the optimization problem can be rewritten as:
\begin{align}
    &\begin{aligned}\min_{\mu_{i+1}, \Theta_{i+1}} \ -\frac{1}{K}\sum_{k=1}^{K}  V_{\omega_i}(s_0^k, c^k)\sqrt{\frac{\left|\Theta_{i+1}\right|}{\left|\Theta_i\right|}}\exp
       \Big[& -\frac{1}{2} \Vert \mu_{i+1} - c^k \Vert_{\Theta^{-1/2}_{i+1}\widetilde{\Sigma}^{-1}\Theta^{-1/2}_{i+1}}^2 \\
   & +\frac{1}{2} \Vert \mu_{i} - c^k \Vert_{\Theta^{-1/2}_{i}\widetilde{\Sigma}^{-1}\Theta^{-1/2}_{i}}^2\Big]
    \end{aligned}  \\
    \text{s.t. } &\nonumber\\
    &\begin{aligned}\frac{1}{2}\Big[& \Vert \mu_{i+1} - \mu_{i} \Vert_{\Theta^{-1/2}_{i+1}\widetilde{\Sigma}^{-1}\Theta^{-1/2}_{i+1}}^2 - \ln \frac{|\Theta_{i+1}|}{|\Theta_{i}|} \\ 
       & + \mathrm{tr} \left(\Theta^{-1/2}_{i}\widetilde{\Sigma}^{-1}\Theta^{-1/2}_{i} \Theta^{1/2}_{i+1}\widetilde{\Sigma}\Theta^{1/2}_{i+1}  \right) - d\Big]\leq \epsilon
    \end{aligned} \\
    & \Theta_{i+1} >0
\end{align}
This optimization problem has a closed solution if solved in a block coordinate descent separating $\theta$ and $\mu$. Also we need to further simplify it by using Taylor expansion of objective function and constraints. It must be noted that the approximations are going to be accurate as required if $\epsilon$ is picked small enough. 

\textbf{Lemma 1.} \textit{Given the assumptions the performance optimization problem has a unique solution}
\begin{align}
   H_i =& \frac{1}{2}\Theta_i^{-1}( \mathrm{I}+\widetilde{\Sigma} \circ \widetilde{\Sigma}^{-1})\Theta_i^{-1}\\
   \omega_i =& \frac{1}{2}\left[(\Theta_i^{-2}  - \Theta_i^{-3/2} (\widetilde{\Sigma} \circ \widetilde{\Sigma}^{-1})\Theta_i^{-3/2} ) \bm{\theta}_i - (\Theta_i^{-3/2}\InvSigma \Theta_i^{-1/2}(\widetilde{\mu} - \mu_i))\circ(\widetilde{\mu}- \mu_i)\right]\\
   \overline{u}_i =& \frac{1}{K}\sum_{k=1}^K V_{\omega_i}(s_0^k, c^k)(c^k - \mu_i )\\
    \overline{V}_i =& \frac{1}{K}\sum_{k=1}^K V_{\omega_i}(s_0^k, c^k)\\
    \overline{\psi}_i = & \frac{1}{2K}\sum_{k=1}^K V_{\omega_i}(s_0^k, c^k)\biggl(\left( (\Theta_i^{-3/2}\widetilde{\Sigma}^{-1}\Theta_i^{-1/2}) (c^k - \mu_i)\right)\circ (c^k - \mu_i)  -\Theta_i^{-2} \bm{\theta}_i\bigg) \\
    \mu_{i+1}^\star = &\mu_{i} + \sqrt{2\varepsilon} \frac{\ubar_i}{\Wnorm{\ubar_i}{\Sigma_i^{-1}}}\label{eq:opt1mu}\\
    \bm{\theta}_{i+1}^\star =&  \bm{\theta}_i + 2\sqrt{\varepsilon}\frac{H^{-1} \overline{\psi}_i}{\Vert  \overline{\psi}_i\Vert_{H^{-1}}}
\end{align}

Lemma 1. update rules are intuitive, Equation~\ref{eq:opt1mu} implies that the mean of the context has to shift towards a weighted average of the context sampled where these weights are proportional to the expected reward that the agent collected. Furthermore, the $\theta$ update rule tries to balance covariance matching the shift in the $\mu$. Given a reasonable $V_-$ the agent learns a meaningful policy

\subsection{Context Distribution Convergence Problem}
Following the same  simplifications done in the Performance optimization problem the convergence problem then can be reformulated as:

\begin{align*}
  \ &\min_{\mu_{i+1}, \Theta_{i+1}} \ \frac{1}{2}\left[ \Wnorm{\Theta_{i+1}^{-1/2}(\mu_{i+1}- \widetilde{\mu})}{\InvSigma}^2 +\mathrm{tr}\left(\Theta_{i+1}^{-1/2}\InvSigma\Theta_{i+1}^{-1/2} \widetilde{\Sigma}\right)  + \ln |\Theta_{i+1}| - d \right] \\
    \text{s.t. } &\\
    &\begin{aligned}   
    \frac{1}{K}\sum_{k=1}^{K}  V_{\omega_i}(s_0^k, c^k)\sqrt{\frac{|\Theta_{i+1}|}{|\Theta_i|}}\exp{}-\frac{1}{2}\bigg[&\Wnorm{\Theta_{i+1}^{-1/2}(c^k-\mu_{i+1})}{\InvSigma}^2  \\
    &- \Wnorm{\Theta_{i}^{-1/2}(c^k-\mu_i)}{\InvSigma}^2\bigg] \geq V_{-}\end{aligned}\\
    &\frac{1}{2}\Big[ \Vert \mu_{i+1} - \mu_{i} \Vert_{\Theta^{-1/2}_{i+1}\widetilde{\Sigma}^{-1}\Theta^{-1/2}_{i+1}}^2 - \ln \frac{|\Theta_{i+1}|}{|\Theta_{i}|}         + \mathrm{tr} \left(\Sigma_i^{-1}\Theta^{1/2}_{i+1}\widetilde{\Sigma}\Theta^{1/2}_{i+1}  \right) - d\Big]\leq \epsilon\\
    & \Theta_{i+1} >0
\end{align*}
The optimization problem is similarly made linear by Taylor expansion then analytically solved using a block coordinate descent approach.

\textbf{Lemma 2.} \textit{Given the assumptions the Context convergence problem has a unique solution}
\begin{align}
    \begin{bmatrix}
    \lambda_1^\star\\
    \lambda_2^\star
    \end{bmatrix} =& \begin{cases}
\begin{bmatrix}
    0\\
    1
\end{bmatrix} \qquad&\text{, \  \ \ } \begin{aligned}
     (\overline{V}_i- V_{-}) + \Wprod{\ubar_i}{\widetilde{\mu} - \mu_i}{\Sigma_i^{-1}} &\geq 0\\
    \Wnorm{\widetilde{\mu} - \mu_i}{\Sigma_i^{-1}} &\geq 2\varepsilon
\end{aligned}\\
     \begin{bmatrix}
     - \frac{\Vbar_i - V_- + \Wprod{\ubar_i}{\widetilde{\mu} - \mu_i}{\Sigma_i^{-1}}}{\Wnorm{\ubar}{\Sigma_i^{-1}}^2}\\
    1
\end{bmatrix}\qquad&\text{, } \begin{aligned} \Vbar_i - V_- + \Wprod{\ubar_i}{\widetilde{\mu} - \mu_i}{\Sigma_i^{-1}} &\leq 0\\ 
     \frac{(\Vbar_i - V_- )^2 - \Wprod{\ubar_i}{\widetilde{\mu}- \mu_i}{\Sigma_i^{-1}}^2}{\Wnorm{\ubar_i}{\Sigma_i^{-1}}^2}&\\
     +\Wnorm{\widetilde{\mu} - \mu_i}{\Sigma_i^{-1}}^2 &\leq 2\varepsilon 
\end{aligned}\\
\begin{bmatrix}
     0\\
    \frac{\Wnorm{\widetilde{\mu}- \mu_i}{\Sigma_i^{-1}}}{\sqrt{2\varepsilon}}
\end{bmatrix}\qquad&\text{, }\quad \begin{aligned}  \Wnorm{\widetilde{\mu} - \mu_i}{\Sigma_i^{-1}}^2 &\geq 2\varepsilon\\ 
     \frac{\Vbar_i - V_-}{\sqrt{2\varepsilon}}+\frac{\Wprod{\ubar_i}{\widetilde{\mu}- \mu_i}{\Sigma_i^{-1}}^2}{\Wnorm{\widetilde{\mu} - \mu_i}{\Sigma_i^{-1}}} &\leq 0
\end{aligned}\\
\begin{bmatrix}
     -\frac{(\Vbar_i - V_-) \lambda_2^\star -   \Wprod{\ubar_i}{\widetilde{\mu}- \mu_i}{\Sigma_i^{-1}}}{\Wnorm{\ubar}{\Sigma_i^{-1}}^2}\\
   \sqrt{ \frac{\Wnorm{\widetilde{\mu}- \mu_i}{\Sigma_i^{-1}}^2 \Wnorm{\ubar_i}{\Sigma_i^{-1}}^2 - \Wprod{\ubar_i}{\widetilde{\mu}- \mu_i}{\Sigma_i^{-1}}^2 }{2\varepsilon \Wnorm{\ubar_i}{\Sigma_i^{-1}}^2 - (\Vbar_i - V_-)^2}}
\end{bmatrix}\qquad&\text{, } \text{ Otherwise}
\end{cases}\\
\mu_{i+1} =& \mu_i + \frac{1}{\lambda^\star_2}(\widetilde{\mu} - \mu_i +\lambda^\star_2 \ubar)\\
\begin{bmatrix}
    \lambda_3^\star\\
    \lambda_4^\star
    \end{bmatrix} =& \begin{cases}
\begin{bmatrix}
    0\\
    \frac{\Wnorm{\omega_i}{}}{2\sqrt{\varepsilon}}
\end{bmatrix} \qquad&\text{, \  \ \ } \Vbar_i - V_- - \frac{2\sqrt{\varepsilon}}{\Wnorm{\omega_i}{}}\Wprod{\psi_i}{\omega_i}{H_i^{-1}}\geq 0\\[10pt]
     \begin{bmatrix}
     \frac{\Wprod{\psibar_i}{\omega_i}{H_i^{-1}}- \lambda_4^\star (\Vbar_i - V_-)}{\Wnorm{\psibar_i}{H_i^{-1}}}\\
     \Bigl(\frac{\Wnorm{\omega_i}{}^2 \Wnorm{\psibar_i}{}^2 - \Wprod{\omega_i}{\psibar_i}{}^2}{4\varepsilon \Wnorm{\psibar_i}{}^2 - (\Vbar - V_-)^2}\Bigr)^{0.5}
\end{bmatrix}\qquad&\text{, \ \ \ Otherwise}\\  
\end{cases}\\
\bm{\theta}_{i+1} = & \begin{cases}\bm{1}\qquad &\text{, \ }  \begin{aligned}\Vbar_i - V_- + \Wprod{\psibar_i}{\bm{1} - \bm{\theta_i}}{} &\geq 0\\
\Wnorm{\bm{1} - \bm{\theta_i}}{H_i}^2 &\leq 4\varepsilon
\end{aligned}\\
\bm{\theta}_i + \frac{1}{\lambda_4^\star}H_i^{-1}(\lambda_3^\star\psibar_i - \omega_i)\qquad &\text{, \ } \text{Otherwise} \\
\end{cases}
\end{align}
The $\mu$ update implies that the sampling context distribution mean is a weighted average of target context distribution mean, current context distribution mean and the expected best performing context mean. This can be interpreted as an expansion of default curriculum which is equivalent to the unconstrained optimization problem with the that  leads to sampling the target distribution immediately.  
 
\textbf{Corollary to Lemma 2.} \textit{$\theta$ and $\mu$ respectively converge to $\bm{1}$ and $\widetilde\mu$ as $i\rightarrow \infty$, therefore agent context distribution asymptotically  becomes the target context distribution.}

The convergence to the target distribution is critical to any curriculum learning algorithm. Assuming  the optimization problem has a feasible solution it would iteratively converge to the target context distribution. Consequently it would not converge to the context distribution if and only if the performance constraint is not satisfied with the reinforcement learning algorithm. Even in that case the algorithm would do no worse than the default curriculum.

\subsection{Algorithmic Realization}
The Curriculum learning  controls the context space samples. The rest of the learning process is not changed and is determined in the learning algorithm.
\begin{algorithm}
	\caption{Gaussian Self-paced Deep Reinforcement Learning} 
	\begin{algorithmic}[1]
 \State \textbf{Input:} Initial context distribution, policy parameters $\omega_0$, Target context distribution, CL Hyper-parameters
		\For {$i=1,2,\ldots N$}
  \Statex \textbf{Agent update}
				\State Sample contexts: $c^k \sim p_{\nu_i}(c), k = 1, \dots, K$
				\State Rollout trajectories: $\tau^k \sim \pi_{\omega_i}(\tau | c^k), k = 1, \dots K$ $\hat{A}_{1},\ldots,\hat{A}_{T}$
    \State Update policy according to learning algorithm: $\pi_{\omega_{i+1}}$
    \Statex \textbf{Curriculum learning update}
    \If{$i\mod N_{CL} =0$}
    \State Update context sampling distribution using Lemma 1. and 2.
    \EndIf
		\EndFor
	\end{algorithmic} 
\end{algorithm}

\section{Experiments}
This section investigates the performance the proposed curriculum learning algrthim, SPGL, and compares it to the SPDL algorithm. The evaluation is done using the RLLib implementation of PPO  learning algorithm\cite{Liang2017}. The learning algorithm is then interacts with the environment through the SPGL curriculum. The applicabality of SPGL is not limited to PPO algorithm and as shown in \cite{klink2020spdrl}, any learning algorithm would be become more  efficient.

\subsection{Point Mass Environment}

In the point mass environment the objective of the agent is to move a dot through the gate to the target destination. The size and location of the gate and the floor friction is determined by the context resulting in $c \in \R^3$. The state space is the linear location and speed of the point mass. The initial location of the point mass and target location is fixed. The action determines the force applied to the point mass by the agent.  The agent is awarded as based on the distance to the target location. Crashing the point mass into the wall terminates the episode prematurely. The agent is evaluated in two scenarios, Setup 1 with a hidden context variable coming from a narrow gaussian context variable. In Setup 2,  the location of the gate has a high variance hence the context variable is required to be  observable.

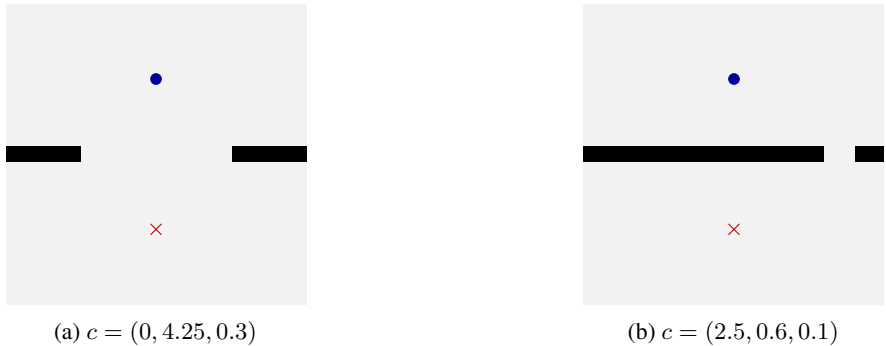
\begin{figure}[h]     
\begin{subfigure}[b]{0.45\textwidth}
         \centering
             \begin{tikzpicture}
\fill[gray!10] (-2,-2) rectangle (2,2);
\filldraw [black!40!blue] (0,1) circle (2pt);
\fill [black] (-2,-0.1) rectangle (2, 0.1);  
\fill [gray!10] (-1,-0.2) rectangle (1, 0.2);  
\draw [color=black!30!red, cap = round] (-0.07,-.93)-- (0.07,-1.07) ;
\draw [color =black!30!red, cap = round] (-0.07,-1.07)-- (0.07,-.93) ;
\end{tikzpicture}

         \caption{$c = (0, 4.25, 0.3)$}
         \label{fig:pm1}
\end{subfigure}
\hfill
\begin{subfigure}[b]{0.45\textwidth}
        \centering
           \begin{tikzpicture}
\fill[gray!10] (-2,-2) rectangle (2,2);
\filldraw [color = black!40!blue] (0,1) circle (2pt);
\fill [black] (-2,-0.1) rectangle (2, 0.1);  
\fill [gray!10] (1.2,-0.2) rectangle (1.6, 0.2);  
\draw [color=black!30!red,cap=round] (-0.07,-.93)-- (0.07,-1.07) ;
\draw [color =black!30!red, cap=round] (-0.07,-1.07)-- (0.07,-.93) ;
\end{tikzpicture}

         \caption{$c = (2.5, 0.6, 0.1)$}
         \label{fig:pm2}
\end{subfigure}
     
  \caption{Point mass environment with initial context sample Figure~\ref{fig:pm1} and target context sample in Figure~\ref{fig:pm2}}
  \label{fig:PM}
\end{figure}

\begin{table}[t]
\caption{Performance summary of CL methods for Point Mass environment(setup 1)}

    \centering
   \begin{tabular}{|c||c|c|c|c|c|}
\hline
Curriculum&\multicolumn{3}{c|}{ Average Collected Reward}& \multicolumn{2}{c|}{Success Rate}\\
\cline{2-6}
 &mean&se&p-value &mean & se\\
\hline
Default&17.09&1.426&\cellcolor{red!25}0.008&37&17.1\\
Self-paced&22.16&0.201&0.071&99&0.7\\
Gaussian Self-paced&\cellcolor{green!25}22.64&\cellcolor{green!25}0.098& &\cellcolor{green!25}100&\cellcolor{green!25}0.0\\
\hline
\end{tabular}
\label{tab:PM2res}
\end{table}

\begin{figure*}[h]
\begin{subfigure}[t]{.45\textwidth}
\centering
\includegraphics{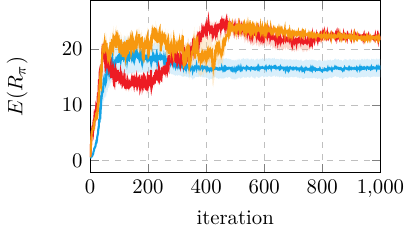}

    \caption{Average Collected Reward during training}
    \label{fig:pm_s1hid_rew}
\end{subfigure}\hfill
\begin{subfigure}[t]{.45\textwidth}
\centering
\includegraphics{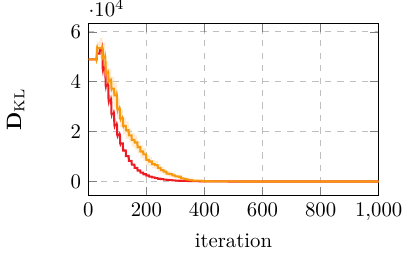}
    \caption{KL-divergence between the sampling context  distribution and target context distribution}
    \label{fig:pm_s1hid_kl}
\end{subfigure}\\
\centering
\begin{tikzpicture} 
    \begin{axis}[%
    hide axis,
    xmin=10,
    xmax=50,
    ymin=0,
    ymax=0.4,
    legend style={draw=black,legend cell align=left, legend columns=-1}
    ]
    \addlegendimage{Cerulean,mark=.}
    \addlegendentry{Default };
    \addlegendimage{Red,mark=.}
    \addlegendentry{Self-paced };
    \addlegendimage{YellowOrange,mark=.}
    \addlegendentry{Gaussian Self-paced };
    \end{axis}
\end{tikzpicture}

\caption{Point Mass setup 1 experiment with hidden context curriculum learning and training}\label{fig:PM_result1}
\end{figure*}
As shown in Table \ref{tab:PM2res}, the SPGL curriculum depicts better robustness compared to SPDL. Both methods however, excel the Default curriculum with statistically significant improvements in terms of average collected rewards and success rate.

\begin{table}[t]
\caption{Performance summary of CL methods for Point Mass environment(setup 2)}

    \centering
     \begin{tabular}{|c||c|c|c|c|c|}
\hline
Curriculum&\multicolumn{3}{c|}{ Average Collected Reward}& \multicolumn{2}{c|}{Success Rate}\\
\cline{2-6}
 &mean&se&p-value &mean & se\\
\hline
Default&17.31&1.438&\cellcolor{red!25}0.011&37.5&17.12\\
Self-paced&\cellcolor{green!25}22.57&\cellcolor{green!25}0.028& &\cellcolor{green!25}99.94&\cellcolor{green!25}0.00\\
Gaussian Self-paced&22.25&0.305&0.359&98.39&1.41\\
\hline
\end{tabular}
\label{tab:PM3res}
\end{table}

\begin{figure}[h]
\begin{subfigure}[t]{.45\textwidth}
\centering
\includegraphics{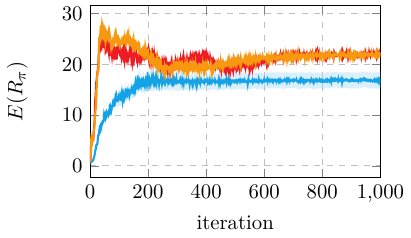}

    \caption{Average Collected Reward during training}
    \label{fig:PM_vis_rew}
\end{subfigure}\hfill
\begin{subfigure}[t]{.45\textwidth}
\centering
\includegraphics{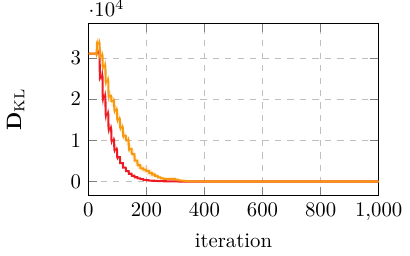}

    \caption{KL-divergence between the sampling context  distribution and target context distribution}
    \label{fig:PM_vis_kl}
\end{subfigure}\\
\begin{tikzpicture} 
    \begin{axis}[%
    hide axis,
    xmin=10,
    xmax=50,
    ymin=0,
    ymax=0.4,
    legend style={draw=black,legend cell align=left, legend columns=-1}
    ]
    \addlegendimage{Cerulean,mark=.}
    \addlegendentry{Default };
    \addlegendimage{Red,mark=.}
    \addlegendentry{Self-paced };
    \addlegendimage{YellowOrange,mark=.}
    \addlegendentry{Gaussian Self-paced };
    \end{axis}
\end{tikzpicture}
\centering
\caption{Point Mass setup 2 experiment with visible context curriculum learning and training}\label{fig:PM_result2}
\end{figure}

The Table~\ref{tab:PM3res}, summarizes the performance comparison of SPGL and the baselines in Setup 2 with visible context. Similar to setup 1 both self paced methods significantly outperformed the default curriculum. The learning curves in Figure~\ref{fig:PM_result2}, are consistent with the evaluations and while both self paced learners have learned a policy that navigates the floor according to the context variable, they have alos converged to the target context distribution.

\subsection{Lunar Lander Environment}

Another test environment used in this study is the Continuous Lunar Lander with a hidden context space. The goal is to optimize the rocket's landing trajectory while balancing performance and robustness. The agent receives rewards for approaching the landing pad and penalties for crashing or excessive fuel consumption.

The observation space is 8-dimensional, including the rocket's position and velocity along the x and y axes, its orientation, and two binary indicators representing ground contact on each landing leg. The action space is continuous, $\mathbb{R}^3$, representing the thrust levels of the main and two side engines.

The context variables which are hidden from the agent include gravity, wind, and turbulence intensity. This setup is designed to evaluate the agent's ability to learn robust policies that generalize across a wide range of environmental conditions. By withholding the context, the task simulates real-world uncertainties and emphasizes adaptability in policy learning.

\begin{figure*}[h]
\begin{subfigure}[t]{.45\textwidth}
\centering
\includegraphics{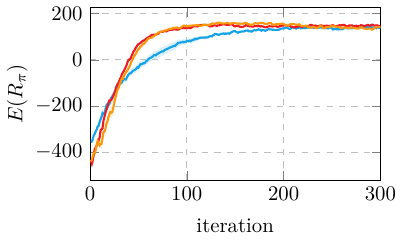}

    \caption{Average Collected Reward during training}
    \label{fig:LLrew}
\end{subfigure}\hfill
\begin{subfigure}[t]{.45\textwidth}
\centering
\includegraphics{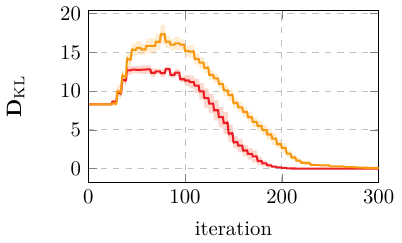}

    \caption{KL-divergence between the sampling context  distribution and target context distribution}
    \label{fig:LLkl}
\end{subfigure}\\
\begin{tikzpicture} 
    \begin{axis}[%
    hide axis,
    xmin=10,
    xmax=50,
    ymin=0,
    ymax=0.4,
    legend style={draw=black,legend cell align=left, legend columns=-1}
    ]
    \addlegendimage{Cerulean,mark=.}
    \addlegendentry{Default };
    \addlegendimage{Red,mark=.}
    \addlegendentry{Self-paced };
    \addlegendimage{YellowOrange,mark=.}
    \addlegendentry{Gaussian Self-paced };
    \end{axis}
\end{tikzpicture}
\centering
\caption{Lunar Lander experiment with hidden context curriculum learning and training}\label{fig:LL_result1}
\end{figure*}

\begin{table}[t]
\caption{Performance summary of CL methods for Lunar Lander environment}

    \centering
    \begin{tabular}{|c||c|c|c|c|c|}
\hline
Curriculum&\multicolumn{3}{c|}{ Average Collected Reward}& \multicolumn{2}{c|}{Success Rate}\\
\cline{2-6}
 &mean&se&p-value &mean & se\\
\hline
Default&246.55&3.458&\cellcolor{red!25}0.042&89&1.3\\
Self-paced&252.59&3.199&0.373&88&1.9\\
Gaussian Self-paced&\cellcolor{green!25}256.29&\cellcolor{green!25}1.921& &\cellcolor{green!25}90&\cellcolor{green!25}0.8\\
\hline
\end{tabular}
\label{tab:LLres}
\end{table}

As shown in Table~\ref{tab:LLres}, the SPGL agent significantly outperforms all other methods, providing strong evidence that the proposed curriculum effectively enhances sample efficiency. While the task is not fully solved, the superior performance of the SPGL agent demonstrates its robustness in learning under sparse rewards. Furthermore, Table~\ref{fig:LL_result1} highlights a key advantage of SPGL over SPDL: the context optimization process in SPGL avoids the numerical instabilities inherent in SPDL. The use of a closed form solution in SPGL offers greater stability and is inherently less susceptible to the randomness typically encountered in numerical optimization procedures.

\subsection{Ball Catching Environment}
The ball-catching environment simulates a Barrett WAM robotic arm tasked with intercepting a ball thrown toward it\citep{klink2020spdrl}. The agent controls the torques applied to the robot's joints. The arm is constrained to move within a plane that intersects the ball's trajectory at a single point. The coordinates where the ball reaches the catching plane, as well as the initial distance from the arm, are determined by the context variable, hence remains unobservable to the agent.

The reward function is sparse, consisting of a penalty for excessive movement and a positive reward only when the ball is successfully caught. While the agent consistently learns to minimize movement to reduce penalties, solving the task through random exploration is nearly impossible. However, the initial context distribution can be set as the initial position of the arm guaranteeing successful catches. 

\begin{figure*}[h]
\begin{subfigure}[t]{.45\textwidth}
\centering
\includegraphics{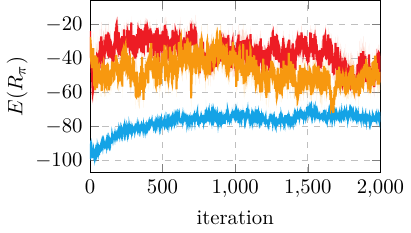}

    \caption{Average Collected Reward during training}
    \label{fig:BCrew}
\end{subfigure}\hfill
\begin{subfigure}[t]{.45\textwidth}
\centering
\includegraphics{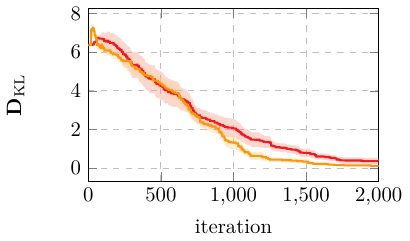}

    \caption{KL-divergence between the sampling context  distribution and target context distribution}
    \label{fig:BCkl}
\end{subfigure}\\
\begin{tikzpicture} 
    \begin{axis}[%
    hide axis,
    xmin=10,
    xmax=50,
    ymin=0,
    ymax=0.4,
    legend style={draw=black,legend cell align=left, legend columns=-1}
    ]
    \addlegendimage{Cerulean,mark=.}
    \addlegendentry{Default };
    \addlegendimage{Red,mark=.}
    \addlegendentry{Self-paced };
    \addlegendimage{YellowOrange,mark=.}
    \addlegendentry{Gaussian Self-paced };
    \end{axis}
\end{tikzpicture}
\centering
\caption{Ball Catching experiment with hidden context curriculum learning and training}\label{fig:BC_res}
\end{figure*}

As shown in Figure~\ref{fig:BCkl}, self-paced agents  converge to the target context distribution. The average collected reward, as well as the success rate consistently remain higher than the agent trained on the default curriculum.

The evaluation results in Table~\ref{tab:BCres} show that the SPGL  significantly outperforms the rest.

\begin{table}[t]
\caption{Performance summary of CL methods for Ball Catching environment}

    \centering
    \begin{tabular}{|c||c|c|c|c|c|}
\hline
Curriculum&\multicolumn{3}{c|}{ Average Collected Reward}& \multicolumn{2}{c|}{Success Rate}\\
\cline{2-6}
 &mean&se&p-value &mean & se\\
\hline
Default&-60.96&4.187&\cellcolor{red!25}0.0&7&2\\
Self-paced&-51.11&6.524&\cellcolor{red!25}0.009&21&2.5\\
Gaussian Self-paced&\cellcolor{green!25}-24.56&\cellcolor{green!25}4.662& &\cellcolor{green!25}31&\cellcolor{green!25}2\\
\hline
\end{tabular}
\label{tab:BCres}
\end{table}

\section{Conclusion}

This work introduced Self-Paced Gaussian Curriculum Learning, a curriculum  learning algorithm that leverages a closed form Gaussian parameterization of the context distribution in contextual MDPs. Building on the self paced learning framework, SPGL decomposes the curriculum update into a performance optimization step and a context convergence step, both of which admit analytical updates for the mean and covariance scaling parameters under mild assumptions. This structure preserves the key advantages of SPDL focus on easy yet informative contexts early in training and convergence to a predefined target distribution, while substantially reducing the optimization overhead typically associated with self-paced methods.

Compared to SPDL, SPGL eliminates the need for repeated numerical inner loop context distribution optimization. By exploiting the closed form KL divergence between multivariate Gaussian distributions and adopting Taylor-based approximations in a block coordinate descent scheme, SPGL replaces unstable or costly numerical solvers with deterministic update rules derived in Lemma~1 and Lemma~2. As a result, the curriculum update becomes both computationally cheaper and numerically more stable. This reduced complexity is particularly relevant when curriculum updates occur frequently or when the context space is moderately high dimensional, where SPDL's iterative optimization can become the dominant computational bottleneck.

The empirical evaluation on three CMDP benchmarks Point Mass, Continuous Lunar Lander, and Ball Catching environments demonstrates the effectiveness and robustness of the proposed approach. In all environments, SPGL matches or outperforms SPDL and the default baseline in terms of average collected reward and success rate, while maintaining stable and monotonic convergence of the sampling context distribution toward the target distribution. The Point Mass and Lunar Lander experiments highlight that SPGL can exploit both hidden and visible context structures, achieving higher performance with statistically significant improvements over default training. In the Ball Catching environment, characterized by extremely sparse rewards and high exploration difficulty, SPGL yields the largest improvements, substantially increasing both average return and success rate relative to SPDL and the default curriculum.

These results suggest that the combination of self-paced curriculum principles with a Gaussian parametric form strikes a favorable balance between expressivity and tractability. The closed form nature of SPGLs updates avoids the numerical instabilities and randomness that can arise from iterative inner-loop optimization in SPDL, leading to more predictable training dynamics and reduced sensitivity to hyperparameters governing the curriculum update. At the same time, the algorithm retains the essential property that the agent's sampling distribution converges to the target context distribution, ensuring that final performance is evaluated on the desired task family rather than an auxiliary or simplified subset.

Given its improved stability, lower computational complexity, and strong performance in sparse reward and partially observed settings, SPGL appears particularly promising for real world applications where environment interactions are expensive and context spaces are structured but not too high-dimensional. Future work includes extending the proposed framework beyond unimodal Gaussian curricula, for example through mixtures of Gaussians or other structured families, and integrating SPGL with a broader class of deep RL algorithms and domains, such as multi agent systems. Another promising direction is to study theoretical guarantees for convergence rates and sample complexity under SPGL, further clarifying the regimes in which Gaussian self paced curricula provide the largest benefits over default training and more complex self paced baselines.

\bibliographystyle{plain}
\bibliography{references.bib}
\appendix
\section{Lemma 1. Proof}
Assuming a coordinate block descent optimization, at each optimization step either $\theta$ or $\mu$ is being updated, creating two independent optimization problems. Additionally the kl-divergence constraint simplifies to a proximity constraint, enabling polynomial approximation of the optimization problem by application of Taylor Expansion. For simplicity without loss of generality here the initial values are represented by $\mu_0$ and $\theta_0$ and the updated values are represented by $\mu$ and $\theta$.

\begin{enumerate}[I.]
 \item $\mu$ update rule\\
    \begin{align*}
      &\EXP (V_\omega| \mu, \Theta_0) = -\frac{1}{K}\sum_{k=1}^{K}\left.  V_{\omega_0}(s_0^k, c^k)\exp{-\frac{1}{2}\left[\Wnorm{\mu- c^k}{\Sigma_0^{-1}}^2 - \Wnorm{\mu_0- c^k}{\Sigma_0^{-1}}^2 \right]}\right. \\
   & \qquad \qquad \quad \approx \Vbar + \Wprod{\ubar}{\mu - \mu_0}{\Sigma_0^{-1}} \\
       & \mathbf{D}_\text{KL}\left(\mathcal{N}(\mu, \Sigma_0)\Vert \mathcal{N}(\mu_0, \Sigma_0)\right) =\frac{1}{2} \Wnorm{\mu - \mu_0}{\Sigma_0^{-1}}^2\\
     \end{align*}
     Then optimization problem is equivalent to:
    \begin{align*}
    &\min_{\mu} \  -  \Wprod{\ubar}{\mu - \mu_0}{\Sigma_0^{-1}} \\
    \text{s.t. } &  \frac{1}{2} \Wnorm{\mu - \mu_0}{\Sigma_0^{-1}}^2\leq \varepsilon\\
    \end{align*}
    The Lagrangian then  is:
    \begin{align*}
    \mathcal{L}(\mu, \lambda) &=   - \Wprod{\ubar}{\mu - \mu_0}{\Sigma_0^{-1}} + \lambda \left[\frac{1}{2} \Wnorm{\mu - \mu_0}{\Sigma_0^{-1}}^2 -\varepsilon \right]\\
    \Rightarrow &\begin{cases}\frac{\partial\, \mathcal{L}}{\partial \, \mu} = -\Sigma_0^{-1}\ubar_0 + \lambda \Sigma_0^{-1} (\mu - \mu_0) = 0&\\
    \frac{\partial\, \mathcal{L}}{\partial \, \lambda} = \frac{1}{2} \Wnorm{\mu - \mu_0}{\Sigma_0^{-1}}^2  -\varepsilon = 0 &\text{ or } \lambda = 0
    \end{cases}\\
    \Rightarrow &\begin{cases} \mu^\star = \mu_0 +\frac{1}{\lambda }\ubar_0 \\
    \lambda^\star = \frac{\Vert \ubar_0\Vert_{\Sigma_0^{-1}}}{\sqrt{2\varepsilon}}\geq 0
    \end{cases}\\
    \end{align*}
    Since the loss function is linear with respect to 
    $\mu$, the K.K.T. conditions are satisfied.
    
    \item $\Theta$ update rule\\

    \begin{align*}
 &\EXP (V_\omega| \mu_0, \Theta) = -\frac{1}{K}\sum_{k=1}^{K}\left.  V_{\omega_0}(s_0^k, c^k)\sqrt{\frac{|\Theta_0|}{|\Theta|}}\exp{-\frac{1}{2}\left[\Wnorm{\mu_0- c^k}{(\Theta^{1/2}\widetilde{\Sigma}\Theta^{1/2})^{-1} - \Sigma_0^{-1}}^2 \right]}\right. \\
   & \qquad \qquad \quad \approx \Vbar_0 + \Wprod{\psibar_0}{\bm{\theta} -\bm{\theta}_0}{} \\
       & \mathbf{D}_\text{KL}\left(\mathcal{N}(\mu_0, \Theta^{1/2}\widetilde{\Sigma}\Theta^{1/2})\Vert \mathcal{N}(\mu_0, \Sigma_0)\right) =\frac{1}{2} \left( \text{tr}\left( \Sigma_0^{-1} \Theta^{1/2} \InvSigma\Theta^{1/2}\right) - \ln |\Theta| + \ln |\Theta_0| - d\right)\\
   & \qquad \qquad \quad \approx \frac{1}{4} \Wnorm{\bm{\theta} -\bm{\theta}_0}{H_0}^2 \\
       \end{align*}
     Then optimization problem is equivalent to:
    \begin{align*}
    &\min_{\bm{\theta}} \  -  \Wprod{\psibar_0}{\bm{\theta} - \bm{\theta}_0}{} \\
    \text{s.t. } &  \frac{1}{4} \Wnorm{\bm{\theta} -\bm{\theta}_0}{H_0}^2\leq \varepsilon\\
    \end{align*}
    The Lagrangian then  is:
    \begin{align*}
    \mathcal{L}(\bm{\theta}, \lambda) &=    -\Wprod{\psibar_0}{\bm{\theta} -\bm{\theta}_0}{} + \lambda \left[\frac{1}{4} \Wnorm{\bm{\theta} -\bm{\theta}_0}{H_0}^2 -\varepsilon \right]\\
    \Rightarrow &\begin{cases}\frac{\partial\, \mathcal{L}}{\partial \, \bm{\theta}} =  - \psibar +\frac{\lambda}{2} H_0 (\bm{\theta} - \bm{\theta}_0)= 0&\\
    \frac{\partial\, \mathcal{L}}{\partial \, \lambda} = \frac{1}{4} \Wnorm{\bm{\theta} - \bm{\theta}_0}{H_0}^2  -\varepsilon = 0 &\text{ or } \lambda = 0
    \end{cases}\\
    \Rightarrow &\begin{cases} \bm{\theta}^\star = \bm{\theta}_0 +\frac{1}{2\lambda } H_0^{-1} \psibar_0 \\
    \lambda^\star = \frac{\Wnorm{\psibar}{H^{-1}_0}}{4\sqrt{\varepsilon}}
    \end{cases}
    \end{align*}
    Since the loss function is linear, and both the $\lambda$ and 
    $\bm{\theta}$ are feasible, and the optimization problem is convex the K.K.T. conditions are satisfied.
\end{enumerate} 
\section{Lemma 2. Proof}
After passing the minimum performance constraint the curriculum learning needs to solve the target context distribution convergence optimization problem. Similarly for simplicity the $i+1$ subscripts in $\theta$ and $\mu$ are dropped and $i$ subscripts are changed to 0.

\begin{enumerate}[I.]
    \item $\mu$ update rule\\

\begin{align*}
       & \mathbf{D}_\text{KL}\left(\mathcal{N}(\widetilde{\mu}, \widetilde{\Sigma})\Vert \mathcal{N}(\mu, \Sigma_0)\right) =\frac{1}{2} \left[ \Wnorm{\mu - \widetilde{\mu}}{\Sigma_0^{-1}}^2 + \text{tr}(\Sigma_0^{-1}\widetilde{\Sigma}) -\ln \frac{|\widetilde{\Sigma}|}{|\Sigma_0|}  - d\right]\\
     \end{align*}
     Then optimization problem is equivalent to:
    \begin{align*}
    &\min_{\mu} \  \frac{1}{2}  \Wnorm{\mu - \widetilde{\mu}}{\Sigma_0^{-1}}^2 \\
    \text{s.t. } & \Vbar_0 + \Wprod{\ubar_0}{\mu - \mu_0}{\Sigma^{-1}_0} \geq V_-\\
    &  \frac{1}{2} \Wnorm{\mu - \mu_0}{\Sigma_0^{-1}}^2\leq \varepsilon\\
    \end{align*}
    The Lagrangian then  is:
    \begin{align*}
    \mathcal{L}(\mu, \lambda_1, \lambda_2) &=  \frac{1}{2}  \Wnorm{\mu - \widetilde{\mu}}{\Sigma_0^{-1}}^2 +\lambda_1 \left[ V_- -\Vbar_0 - \Wprod{\ubar_0}{\mu - \mu_0}{\Sigma^{-1}_0}  \right] +(\lambda_2-1) \left[\frac{1}{2} \Wnorm{\mu - \mu_0}{\Sigma_0^{-1}}^2 -\varepsilon \right]\\
    \Rightarrow \frac{\partial\, \mathcal{L}}{\partial \, \mu} &= \Sigma_0^{-1}(\mu - \widetilde{\mu}) -\lambda_1 \Sigma_0^{-1} \ubar_0+ (\lambda_2-1) \Sigma_0^{-1} (\mu - \mu_0)= 0\\
    \Rightarrow \mu^\star &=\mu_0 +\frac{1}{\lambda_2}(\widetilde{\mu} - \mu_0 + \lambda_1 \ubar_0), \qquad \lambda_1>0, \lambda_2>1\\
    \end{align*}
    Given the general solution for the optimal $\mu$ it is only necessary to find bounds for $\lambda_1$ and $\lambda_2$ where their constraints are active, then solve for their values.  
\begin{itemize}
    \item Both constraints are inactive:
\begin{align*}
    \mu^\star = \widetilde{\mu} \Rightarrow \frac{1}{2}\Wnorm{\widetilde{\mu} - \mu_0}{\Sigma_0^{-1}}\leq \varepsilon, \Vbar_0 +\Wprod{\ubar_0}{\widetilde{\mu} - \mu_0}{\Sigma_0^{-1}} \geq V_- 
\end{align*}
\item Proximity constraint is inactive:
\begin{align*}
    &\begin{cases}
        \mu^\star &= \widetilde{\mu} + \lambda_1^\star \ubar_0\\
        \lambda_1^\star  &= -\frac{\Vbar_0 - V_- + \Wprod{\ubar_0}{\widetilde{\mu} - \mu_0 }{\Sigma_0^{-1}}}{\Wnorm{\ubar}{\Sigma_0^{-1}}^2}\\
    \end{cases}\\
    \text{if } & |\lambda_1 + \frac{\Wprod{\ubar_0}{\widetilde{\mu}-\mu_0}{\Sigma_0^{-1}}}{\Wnorm{\ubar_0}{\Sigma_0^{-1}}^2}|\leq \frac{\sqrt{\Wprod{\ubar_0}{\widetilde{\mu}-\mu_0}{\Sigma_0^{-1}}^2 - \Wnorm{\ubar_0}{\Sigma_0^{-1}}^2(\Wnorm{\widetilde{\mu}- \mu_0}{\Sigma_0^{-1}}^2 - 2\varepsilon)}}{\Wnorm{\ubar_0}{\Sigma_0^{-1}}^2}
\end{align*}
\item Performance constraint is inactive:
\begin{align*}
    &\begin{cases}
        \mu^\star &=\mu_0 + \frac{1}{\lambda_2^\star}( \widetilde{\mu} -\mu_0)\\
        \lambda_2^\star  &= \frac{ \Wnorm{\widetilde{\mu} - \mu_0 }{\Sigma_0^{-1}}}{\sqrt{2\varepsilon}}\\
    \end{cases}\\
    \text{if } & \lambda_2^\star \geq - \frac{\Wprod{\ubar_0}{\widetilde{\mu} - \mu_0}{\Sigma_0^{-1}}}{\Vbar - V_-}
\end{align*}
\item Both constraints are active:
\begin{align*}
    &\begin{cases}
        \mu^\star &=\mu_0 + \frac{1}{\lambda_2^\star}( \widetilde{\mu} -\mu_0 + \lambda_1^\star \ubar_0)\\
        \lambda_1^\star  &= -\frac{ \lambda_2^\star(\Vbar_0 - V_-)+\Wprod{\ubar_0}{\widetilde{\mu} - \mu_0 }{\Sigma_0^{-1}}}{\Wnorm{\ubar_0}{\Sigma_0^{-1}}^2}\\
        \lambda_2^\star  &= \sqrt{\frac{ \Wnorm{\widetilde{\mu}- \mu_0}{\Sigma_0^{-1}}^2\Wnorm{\ubar_0}{\Sigma_0^{-1}}^2-\Wprod{\ubar_0}{\widetilde{\mu} - \mu_0 }{\Sigma_0^{-1}}^2}{\Wnorm{\ubar_0}{\Sigma_0^{-1}}^2}}\\
    \end{cases}\\
\end{align*}
    
\end{itemize}
    \item $\Theta$ update rule\\
Linearization of the objective function via Taylor expansion around $\Theta_0$ would yield:
\begin{align*}
        \mathbf{D}_\text{KL}\left(\mathcal{N}(\widetilde{\mu}, \widetilde{\Sigma})\Vert \mathcal{N}(\mu_0, \Theta^{1/2}\widetilde{\Sigma}\Theta^{1/2})\right) &=\frac{1}{2} \left[ \Wnorm{\mu_0 - \widetilde{\mu}}{(\Theta^{1/2}\widetilde{\Sigma}\Theta^{1/2})^{-1}}^2 + \text{tr}((\Theta^{1/2}\widetilde{\Sigma}\Theta^{1/2})^{-1}\widetilde{\Sigma}) -\ln |\Theta| - d\right]\\
        & \approx \mathbf{D}_{\text{KL}}(\left(\mathcal{N}(\widetilde{\mu}, \widetilde{\Sigma})\Vert \mathcal{N}(\mu_0, \Sigma_0)\right) + \Wprod{\omega_0}{\bm{\theta} - \bm{\theta}_0}{}\\
     \end{align*}
 
Given the linearizations given before the optimization problem simplifies to:
\begin{align*}
& \min_{ \bm{\theta}} \ \Wprod{\omega_0}{\bm{\theta} - \bm{\theta}_0}{} \\
    \text{s.t. } & \Vbar+\Wprod{\psibar_0}{\bm{\theta} - \bm{\theta}_0}{} \geq V_-\\
    &  \frac{1}{4} \Wnorm{\bm{\theta} -\bm{\theta}_0}{H_0}^2\leq \varepsilon\\
    \end{align*}
    The Lagrangian then  is:
    \begin{align*}
    \mathcal{L}(\bm{\theta}, \lambda_3, \lambda_4) &=  \Wprod{\omega_0}{\bm{\theta} - \bm{\theta}_0}{}  -\lambda_3(\Vbar - V_- +\Wprod{\psibar_0}{\bm{\theta} -\bm{\theta}_0}{}) + \lambda_4 (\frac{1}{4} \Wnorm{\bm{\theta} -\bm{\theta}_0}{H_0}^2 -\varepsilon )\\
    \Rightarrow \frac{\partial\, \mathcal{L}}{\partial \, \bm{\theta}} &=  \omega_0-\lambda_3  \psibar_0 +\frac{\lambda_4}{2} H_0 (\bm{\theta} - \bm{\theta}_0)= 0\\
    \Rightarrow \bm{\theta}^\star &=\bm{\theta}_0 + \frac{1}{\lambda_4} H_0^{-1}(\lambda_3 \psibar_0 - \omega_0)\\
    \end{align*}

It must be noted that the unconstrained solution is $\bm{\theta} = \mathbf{1}$, which  realizes  in case both constraints are inactive. Therefore the convergence optimization problem has a unique solution as follows:

      \begin{itemize}
    \item Both constraints are inactive:
\begin{align*}
    \bm{\theta}^\star = \bm{1}\text{ if } \frac{1}{4}\Wnorm{\bm{1} - \bm{\theta}_0}{H_0^{-1}}\leq \varepsilon, \Vbar_0 +\Wprod{\psibar_0}{\bm{1} - \bm{\theta}_0}{} \geq V_- 
\end{align*}

\item Performance constraint is inactive:
\begin{align*}
&\begin{cases}
        \bm{\theta}^\star &=\bm{\theta}_0 - \frac{1}{\lambda_4^\star}H_0^{-1}\omega_0\\
        \lambda_4^\star  &= \frac{ \Wnorm{\omega_0}{H_0^{-1}}}{2\sqrt{\varepsilon}}\\   
\end{cases}\\
        \text{if } & \lambda_4^\star \geq  \frac{\Wprod{\psibar_0}{\omega}{H_0^{-1}}}{\Vbar_0 - V_-}
\end{align*}
\item Both constraints are active:
\begin{align*}
    &\begin{cases}
        \lambda_4^\star  &= \bigg(\frac{ \Wnorm{\omega_0}{H_0^{-1}}^2\Wnorm{\psibar_0}{H_0^{-1}}^2-\Wprod{\psibar_0}{\omega_0 }{H_0^{-1}}^2}{4\varepsilon\Wnorm{\psibar_0}{H_0^{-1}}^2- (\Vbar_0 - V_-)^2} \bigg)^{0.5}\\
        \lambda_3^\star & = \frac{\Wprod{\psibar_0}{\omega_0}{H_0^{-1}} - \lambda_4^\star (\Vbar_0 - V_-)}{\Wnorm{\psibar_0}{H_0^{-1}}^2}\\
    \end{cases}\\
\end{align*}
    
\end{itemize}
\end{enumerate}

\section{Corollary to Lemma 2. Proof}
Clearly given the objective function for $\mu$ update step the for any $\Sigma_0$ the $\Wnorm{\mu - \widetilde{\mu}}{\Sigma_0^{-1}}$ would decrease and $\mu$ converges to $\widetilde{\mu}$. 
Similarly, for $\bm{\theta}$,    as  $i\rightarrow \infty$, $\mu_{i+1} = \mu_i$ so both conditions in the convergence optimization problem are relaxed leading to $\bm{\theta} = \bm{1}$.

\section{Experiment Parameters}
The mean and variances used for environments are summarized in Table~\ref{tab:exps}.
\begin{table}[t]
\caption{Experiment parameters summary}
\small{
    \centering
    \begin{tabular}{|c||c|c|c|}
\hline

 Experiment&Target Context Distribution& Initial Context Distribution&$V_-$\\
 \hline
 \multirow{2}{*}{Point Mass(setup 1)}& $\widetilde{\mu}=[2.6, 0.7, 0.1]$&$\mu_0=[0, 4,  2]$&\multirow{2}{*}{5}\\
 &$\widetilde{\theta}=10^{-4}[9, 4, 1]$&$\theta_0=[4, 3.5, 1]$&\\
 \hline
 \multirow{2}{*}{Point Mass(setup 2)}& $\widetilde{\mu}=[2.5, 0.7, 0.1]$&$\mu_0=[0, 4, 2]$&\multirow{2}{*}{5}\\
 &$\widetilde{\theta}=10^{-4}[10000, 9, 1]$&$\theta_0=[4, 3.5, 1]$&\\
 \hline
\multirow{2}{*}{Lunar Lander}& $\widetilde{\mu}=[-10, 15, 1.5]$&$\mu_0=[-7, 10, 1]$&\multirow{2}{*}{1}\\
 &$\widetilde{\theta}=[1, 9, 0.09]$&$\theta_0=[4, 16, 0.16]$&\\
 \hline
 \multirow{2}{*}{Ball Catching}& $\widetilde{\mu}=[0.34\pi, 0.85, 2.37]$&$\mu_0=[0.68, 0.65, 0.8]$&\multirow{2}{*}{42.5}\\
 &$\widetilde{\theta}=[0.1\pi, 0.15, 1]$&$\theta_0=10^{-3}[1, 1, 100]$&\\
 \hline
\end{tabular}}
\label{tab:exps}
\end{table}

\end{document}